\documentclass[11pt]{article}
\usepackage[utf8]{inputenc}
\usepackage[T1]{fontenc}
\usepackage{lmodern}
\usepackage[margin=1in]{geometry}
\usepackage{amsmath,amssymb}
\usepackage{graphicx}
\usepackage{float}
\usepackage{booktabs,array}
\usepackage[table]{xcolor}
\usepackage{caption}
\usepackage{enumitem}
\usepackage{microtype}
\usepackage[colorlinks=true,linkcolor=blue!50!black,citecolor=blue!50!black,urlcolor=blue!50!black]{hyperref}
\title{\Large\bfseries Co-located Mobile AR\\from Brief Mutual Sightings of Users}
\author{Mohammad Rashid\\Department of Mathematics\\The University of Texas at Austin\\\texttt{m.rashid@utexas.edu}}
\date{}
\begin{document}
\maketitle
\begin{abstract}
Co-located augmented reality requires all devices to express virtual content in a common frame. ARKit and ARCore establish this frame by matching visual features of the room, so every device must first scan the same surfaces. In contrast, we propose a registration method based on brief mutual sightings of the users. Since devices align their vertical axis with gravity, two frames differ only by a yaw angle and a horizontal translation. When device A observes the user of device B at bearing $\alpha$ while B observes the user of A at bearing $\beta$, the two lines of sight are antiparallel and the relative yaw is $\alpha - \beta + \pi$, without depth, markers or a commonly observed surface. A sighting is not forced and may last less than a second, which is especially useful in applications where users frequently view each other. Ranged sightings add point correspondences, and a server fits all devices under a Cauchy loss. Devices without depth sensors get dense depth from a point-map network (MoGe-2). In deployment, 85\% of 4,613 iOS devices without LiDAR converged in a median of 24 s (81\% without ultra-wideband ranging), against 87\% of LiDAR devices and 66\% of 3,048 Android phones. Depth arrived within 10 s in 84\% of iOS sessions, when ARKit had found a wall in 32\%. Its scale was within 2\% of ARKit's wall planes but varied by 8\% between frames, which thickened walls about fourfold; an offline per-frame correction reduced wall thickness from 27 to 17 cm on iOS and from 50 to 22 cm on Android. The converged frame still drifted by a median of 6\textdegree{} in yaw. Unlike map sharing, the method needs no scanning step, depth sensor or exchange of images, so a shared session can start on phones as soon as users see each other. The principle also applies to head-worn displays and to robots that observe one another. Future work will run the scale correction on the device and hold the frame once converged.
\end{abstract}

\section{Introduction}

Co-located augmented reality (AR), in which several users in the same room see the same virtual content at the same physical location, requires that all devices express their poses in a common coordinate frame. Each device, however, tracks its own motion in a private frame whose origin is the position at which its session started and whose vertical axis is aligned with gravity. Existing frameworks solve this registration problem visually: ARKit collaborative sessions \cite{r19} and ARCore Cloud Anchors \cite{r20} match image features of the room across devices. These approaches require all devices to observe the same surfaces from similar viewpoints. Consequently, users are typically asked to scan the room before any content can be shared.

In this paper, we propose to register co-located devices from observations of the users themselves instead of the room. Our approach is motivated by a simple observation: in a multi-user session, the cameras are frequently pointed at the other users. Since both frames are aligned with gravity and each device estimates its height above the floor, the frames of two devices differ only by a rotation about the vertical axis and a horizontal translation. If device $A$ observes the user of device $B$ at the same time as $B$ observes the user of $A$, the two lines of sight are antiparallel, and the relative rotation follows directly from the two bearings (Section 3.1). Because this relation does not involve distance, it applies to phones without a depth sensor and before either device has mapped its surroundings. A sighting that additionally carries a range locates the observed user in the frame of the observer. Since the observed device knows its own position in its own frame, each such sighting yields a point correspondence between the two frames. A server aggregates the sightings of all devices and estimates the three unknowns of each device under a Cauchy loss, which limits the influence of erroneous sightings. Note that a sighting is a single camera frame in which a person is detected, and that users are never prompted to look at one another. The frames are merged opportunistically, whenever users come into each other's view.

Shared content must also be consistent with the geometry of the room, since occlusion, collision and placement of virtual objects require dense depth. LiDAR-equipped iPhones obtain it from ARKit scene depth \cite{r1}, and Android phones that support the ARCore Depth API obtain a depth image from it \cite{r2,r3}. However, this support is far from universal. In our deployment, half of the 710 Android models observed, accounting for 15\% of the Android device-sessions, did not support the Depth API, and iPhones without LiDAR have no general rear-camera depth API at all. On these devices, only tracked feature points and planes are available. Both depend on texture and camera motion, so that for the first seconds to minutes of a session the application cannot determine whether a virtual object is behind a wall. To address this limitation, we run MoGe-2 \cite{r4}, a network that predicts a metric point map, on sampled camera frames. The point map leaves a single unknown, a shift along the optical axis, which we recover from the camera intrinsics reported by ARKit or ARCore. The resulting frames are fused into a persistent voxel map, and the scale of each new frame is partially corrected towards overlapping views that are already in the map.

Throughout the paper, we call a system \emph{scan-free} if it provides dense geometry without first asking the user to move the camera around the room, and \emph{shared} if several co-located devices place content in one coordinate frame without exchanging or overlapping visual maps. The system presented here has both properties, and an explicit goal of our evaluation is to determine where it falls short.

We deployed the system in an AR application on iOS and Android and evaluate it on telemetry from this deployment. The telemetry contains the depth grids, camera poses and intrinsics, ARKit or ARCore planes, the sightings and the shared-frame solutions, but neither images nor ground-truth depth. We therefore compare the depth with two references that ARKit and ARCore derive from their visual-inertial metric scale on every device, namely wall planes and camera translation, and use LiDAR devices in the same deployment as a reference population. A small controlled study against LiDAR complements these proxies. Finally, for a sample of sessions, we reprocess the recorded data offline in order to estimate how much of the remaining error could be removed by correcting the scale and the drift of the frame.

In summary, the contributions of this paper are as follows.

\begin{enumerate}[leftmargin=1.6em,itemsep=2pt]
\item We propose a method that registers co-located devices in a common frame from their sightings of each other's users. The relative yaw follows from a single pair of mutual bearings, and the translation from ranged sightings. The method requires neither a shared visual map nor a scanning step.
\item We present a scan-free monocular mapping pipeline for devices without depth sensors, which combines point-map prediction with intrinsics-based shift recovery, frame-to-map scale alignment and visibility-based surface expiry.
\item We report a field evaluation on 4,198 iOS device-sessions from 66 models and 2,864 Android device-sessions from 431 models, with separate analyses of 10,349 devices in multi-device sessions. We quantify how often and how fast the shared frame converges, how the deployed depth compares with ARKit and ARCore geometry and with LiDAR devices, and how much of the error an offline correction removes. The accuracy of the shared frame was not measured against an external reference.
\end{enumerate}

\section{Related Work}

\subsection{Depth on phones}

Valentin et al. \cite{r3} showed that depth from camera motion can drive occlusion and physics on phones, and their method underlies the ARCore Depth API \cite{r2}. DepthLab \cite{r5} organized the uses of such depth into point, surface and dense queries, and our map answers queries of these kinds. These systems rely on parallax, and consequently a stationary or purely rotating camera yields little depth. Dense mobile reconstruction from a single camera, as in MobileFusion \cite{r6}, also requires the user to move the camera through the scene. LiDAR removes this limitation, but only on the devices equipped with it \cite{r1}.

\subsection{Monocular geometry and its scale}

Monocular depth networks differ in the quantities they leave unresolved (Table 1). Relative-depth models such as ZipDepth \cite{r7} predict inverse depth up to an affine transform, which must be fitted to metric references on every frame. Metric depth models such as Depth Anything V2 Metric \cite{r8}, Metric3D \cite{r9}, UniDepth \cite{r10} and Depth Pro \cite{r11} predict metres directly, which makes their scale depend on how well they infer the focal length. MoGe \cite{r12} predicts an affine-invariant point map and recovers its shift from the image geometry, and MoGe-2 \cite{r4} adds a metric scale. With known intrinsics only the shift remains to be solved. The absolute scale, however, remains a learned prior.

\begin{table}[t]
\centering
\small
\captionof{table}{Table 1. Monocular model families and the per-frame parameters they leave unknown.}
\begin{tabular}{>{\raggedright\arraybackslash}p{4.4cm}>{\raggedright\arraybackslash}p{4.3cm}>{\raggedright\arraybackslash}p{2.3cm}>{\raggedright\arraybackslash}p{3.4cm}}
\toprule
\textbf{Model family} & \textbf{Output} & \textbf{Unknown per frame} & \textbf{Resolved by} \\
\midrule
Relative depth (ZipDepth) & inverse depth $q$ up to $w = aq + b$ & $a$, $b$ & fit to metric points \\
Metric depth (Depth Anything V2 Metric) & depth in metres & none nominally & learned prior \\
Metric point map (MoGe-2) & points $(x, y, z)$ and scale $m$ & shift $s$ & camera intrinsics \\
\bottomrule
\end{tabular}
\end{table}

Moreover, per-frame predictions are not consistent over time. Video-depth methods enforce consistency by optimizing over a whole sequence \cite{r13}, which is too slow for a phone. Instead, we fuse each frame into a map online (Section 3.3).

\subsection{Dense mapping}

Volumetric fusion \cite{r14} and voxel maps with ray-cast free-space clearing \cite{r15} are standard for depth sensors. Dense monocular SLAM such as CNN-SLAM \cite{r16} and DROID-SLAM \cite{r17} estimates pose and depth jointly but is too heavy for older phones. In contrast, we take the poses from ARKit's visual-inertial odometry \cite{r18} and fuse the predicted depth into a coarse voxel map.

\subsection{Shared frames}

Existing frameworks for shared AR, such as ARKit collaborative sessions \cite{r19} and ARCore Cloud Anchors \cite{r20}, align devices by matching visual features of the room, which requires the devices to have observed the same surfaces. Collaborative SLAM systems \cite{r21} follow the same principle using shared keyframes. In multi-robot systems, in contrast, the transform between robots is commonly estimated from their observations of each other. Franchi et al. \cite{r22} localize multiple robots from relative measurements whose sender is not identified. Zhou and Roumeliotis \cite{r23} solve the minimal combinations of range and bearing measurements that determine the relative pose of two robots in three dimensions, and Dhiman et al. \cite{r24} recover the relative pose of two cameras from reciprocal observations of fiducial markers mounted on each of them. Our method can be regarded as the planar case of the same constraint, since gravity alignment leaves a single rotation unknown. It differs from these works in what is observed: people detected in ordinary camera frames, without markers, and with a range that is frequently coarse or missing. To the best of our knowledge, the deployment of this approach on phones has not been reported before.

\section{Method}

Figure 1 gives an overview of the system. Each device builds its own map from its own camera, and only camera poses and sightings of other users are transmitted to the server.

\begin{figure}[t]
\centering
\includegraphics[width=0.9\linewidth]{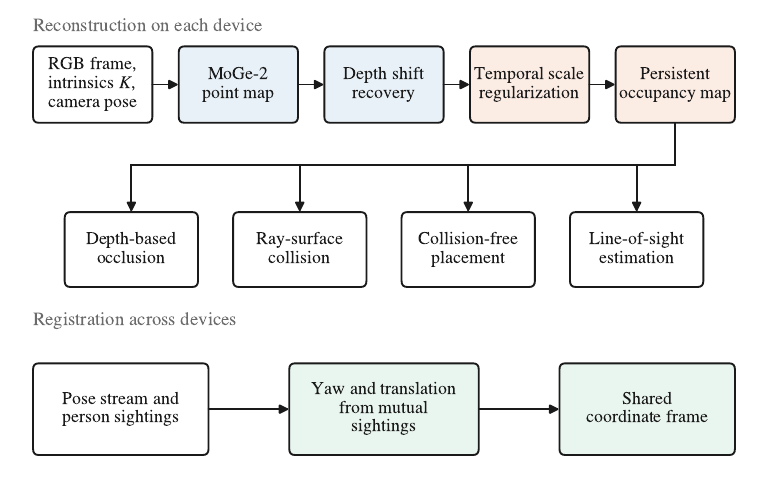}
\captionof{figure}{\textbf{Figure 1.} System architecture. Blue modules recover metric depth from camera frames, orange modules maintain the persistent voxel map and answer queries, and green modules align coordinate frames across devices.}
\end{figure}

\subsection{Shared frame from mutual sightings}

We first describe how co-located devices are registered in a common frame. ARKit and ARCore align the vertical axis of each device's tracking frame with gravity, and each device estimates the height of the floor in its own frame. The frames of two devices therefore differ only by a rotation $\theta$ about the vertical axis and a horizontal translation $\mathbf{t}$. We let one device define the shared frame, so that the problem is to determine these three unknowns for every other device.

\emph{Yaw from a mutual sighting.} Consider two devices $A$ and $B$. Let $\alpha$ denote the horizontal direction in which $A$ observes the user of $B$, expressed in the frame of $A$, and let $\beta$ denote the direction in which $B$ observes the user of $A$ at the same time, expressed in the frame of $B$. Both observations lie on the line connecting the two users, traversed in opposite directions (Figure 2), and hence

\begin{equation*}
\theta = \alpha - \beta + \pi \pmod{2\pi}. \tag{1}
\end{equation*}

Note that Equation (1) involves two bearings and no distance. Since a bearing depends only on the image position of the observed person and on the intrinsics and pose of the camera, the relative yaw of two devices can be determined without depth, without fiducial markers and without the two devices having observed any common part of the room.

\begin{figure}[t]
\centering
\includegraphics[width=0.8\linewidth]{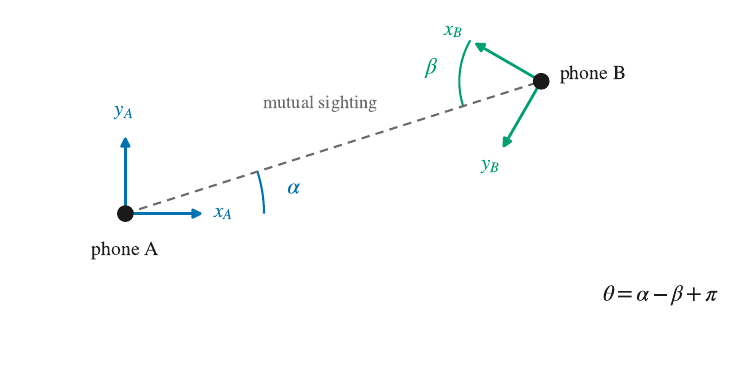}
\captionof{figure}{\textbf{Figure 2.} Relative yaw from mutual sightings. Two lines of sight observed at the same time in two gravity-aligned frames are antiparallel, which fixes the rotation $\theta$ between the frames without any range measurement.}
\end{figure}

\emph{Position from ranged sightings.} A sighting that additionally provides a range locates the observed user at a point in the frame of the observer. Since the observed device knows its own position in its own frame at that time, the sighting constitutes a point correspondence between the two frames. In contrast to Equation (1), such a correspondence does not need to be mutual. Two correspondences at distinct locations, or a single correspondence combined with Equation (1), are sufficient to determine $\theta$ and $\mathbf{t}$.

\emph{Sightings.} Each device runs a person detector on its camera frames, and we define a sighting as a single frame in which a person is detected. A user who is visible to another user for only a fraction of a second therefore already produces a sighting, and the application contains no step that asks users to face each other. The bearing of a detected person is given by the ray through the upper middle of the bounding box, at the torso, projected onto the floor plane. The range is obtained by inverse-variance fusion of up to five estimates: the depth map at the torso, ARKit's people depth, the torso length between pose keypoints, the height of the bounding box under an assumed body height of 1.7 m, and the intersection of the ray towards the feet with the floor. The latter two are only used when the box is tall enough to contain a standing person. Each estimate is assigned a standard deviation that increases with distance, and a single estimate that deviates from the others by more than three standard deviations is discarded. A device transmits at most three sightings per second, each with the fused range $d$ and its standard deviation $\sigma_r$, and samples its own pose every 100 ms.

\emph{Estimation.} The server first estimates the clock offset of each device from message timing. For each sighting, it waits until the poses received from the other devices bracket the time of the sighting and interpolates the position of the observed user, which is assumed to be 0.3 m behind the phone along its heading. A sighting with $\sigma_r \le 0.35\,d$ and $d \le 10$ m contributes a two-dimensional point residual, whereas any other sighting between 0.3 m and 25 m contributes a bearing residual with a standard deviation of $1.5^\circ + \arctan(0.25\,\mathrm{m}/d)$. Mutual sightings that are at most 1.2 s apart are combined according to Equation (1), which is used to initialize the yaw. In addition, three or more such pairs within 30 s that agree to within 12\textdegree{} contribute a relative-yaw residual. The server then minimizes the sum of all residuals over the parameters $(\theta, \mathbf{t})$ of all devices using the Levenberg--Marquardt algorithm with a Cauchy loss whose scale is two standard deviations. This optimization is repeated at most every 0.5 s over the sightings of the last 120 s, and each device adopts a new solution as soon as it is received. The first device to report a pose defines the shared frame.

The estimator accepts three further inputs, all of which are optional. First, the compass provides a prior on the yaw with a standard deviation of at least 25\textdegree{}, which is discarded when mutual sightings contradict it by more than 45\textdegree{}. Second, two iPhones equipped with ultra-wideband radios may contribute the peer position that the radio reports with camera assistance as an additional point. Third, in sessions with three or more users, a sighting is attributed to a user by its agreement with the predicted position of that user once the user has been registered, and before that by the identity tracking of the application or by matching the sighted positions over time against the trajectory of the user.

\emph{Convergence.} From the covariance of the solution, the server predicts the lateral error of a point at a distance of 8 m from the device, $\sigma = (\sigma_t^2 + (8\,\mathrm{m} \cdot \sigma_\theta)^2)^{1/2}$. A device is declared \emph{converged} when $\sigma$ falls below 0.6 m, at least six sightings support the solution (a bearing counts as half a sighting) and the median point residual over 30 s is below 0.6 m. It remains converged as long as both values stay below 0.9 m.

\subsection{Metric depth per frame}

MoGe-2 (ViT-S, 35M parameters) maps a $448 \times 336$ RGB frame to a point map $(x_i, y_i, z_i)$ at $512 \times 384$ and a scale $m$. The point map is correct up to a shift $s$ along the optical axis. With normalized image coordinates $t_{x,i} = (u_i - c_x)/f_x$ and $t_{y,i} = (v_i - c_y)/f_y$ from the intrinsics that ARKit reports for that frame, the shifted points must lie on their pixel rays (Figure 3). We solve

\begin{equation*}
\hat{s} = \arg\min_s \sum_i \left( \frac{x_i}{z_i + s} - t_{x,i} \right)^2 + \left( \frac{y_i}{z_i + s} - t_{y,i} \right)^2 \tag{2}
\end{equation*}

with ten Gauss--Newton steps over an $8 \times 8$ pixel lattice, following MoGe's solver \cite{r12} with the field of view fixed. Depth is then

\begin{equation*}
D_i = m\,(z_i + \hat{s}). \tag{3}
\end{equation*}

Note that Equation (2) determines the shift but not the absolute scale, which is still given by $m$.

\begin{figure}[t]
\centering
\includegraphics[width=0.8\linewidth]{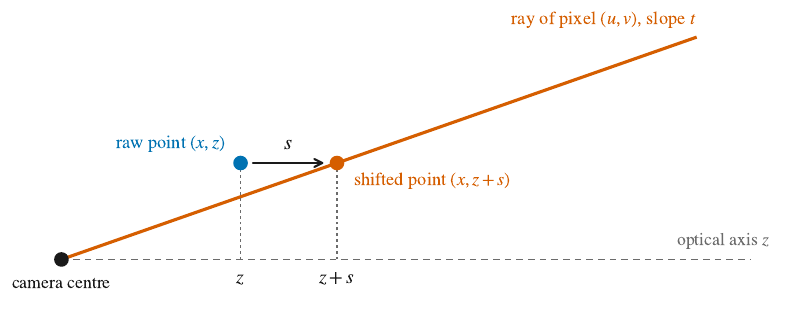}
\captionof{figure}{\textbf{Figure 3.} Shift recovery from known intrinsics. A predicted point off its pixel ray is moved along the optical axis by the shift $s$ until it projects to its pixel.}
\end{figure}

The deployed model uses 8-bit palettized weights (29.7 MB). Its output is resampled to a $252 \times 196$ depth grid, and depth between 0.35 m and 24.9 m is kept. At initialization the model is timed on the CPU and on the Neural Engine, and the Neural Engine is used only if it is at least 15\% faster and its output agrees with the CPU output to within 3\%. Camera frames are sampled for the model at up to 5 Hz in single-device sessions and 3 Hz in multi-device sessions, and less often when the frame budget or the device temperature requires it.

On Android the same checkpoint is converted to LiteRT with a $336 \times 196$ input, the output head stopped at $96 \times 56$, and 8-bit weights, and it runs on the CPU. The shift is recovered from ARCore's intrinsics in the same way.

\subsection{Scale across frames}

A persistent map accumulates every frame, so a frame whose scale is off by a few percent adds a second copy of a wall behind or in front of the first. Before a frame is fused, its points are reprojected into retained views within 1.5 m, and the median ratio $r$ between the frame's depth and those views' depth is computed. The frame is rescaled by

\begin{equation*}
c = \operatorname{clip}\big(1 + \lambda (r - 1),\; 0.92,\; 1.08\big), \quad \lambda = 0.5, \tag{4}
\end{equation*}

with $r$ first divided by its mean over the frame's recent history, so the rule changes relative scale only. Absolute scale therefore stays with the model, apart from two corrections. Pixels that fall within about 18 cm of ARKit's floor height are snapped to the floor. A fit to ARKit feature points can rescale a frame when enough points agree, and in test recordings on an iPhone 11 Pro it rarely did. A stronger setting of Equation (4), $\lambda = 0.8$ and $\pm 12\%$, made free-space placement worse in offline replays of recorded sessions, and was not used.

\subsection{Persistent map}

The map is a voxel grid with 20 cm cells. Each cell stores whether it is occupied, free or unknown, the source of the observation and the time it was observed. A surface seen once expires after 30 s and one seen repeatedly after 120 s. If a newer ray ends behind an occupied cell, the cell is marked as seen through. It then no longer occludes virtual objects or blocks line-of-sight queries, but still blocks moving virtual objects until it expires. A cell observed only once is cleared outright when a later ray passes through it. The map answers four queries, which are occlusion of virtual objects, ray--surface collision, placement in free space and line of sight. Occlusion is resolved per object at the map's 20 cm resolution and not per pixel.

\section{Data and Measures}

\subsection{Deployment telemetry}

Telemetry was collected only from users who declared themselves adults and opted in. Participation is optional and can be withdrawn in the settings. The records used here are depth grids with their camera pose and intrinsics, ARKit or ARCore planes and feature-point counts, frame timing, the sightings and the server's shared-frame solutions. They contain no images, audio or names, identifiers are pseudonymous per session, and we report aggregates only.

We used sessions recorded from 5 to 20 November 2026 on the release described here, read from the telemetry store on 23 November, that lasted at least 60 s (90 s for single-device sessions). We kept every multi-device session with an iPhone or iPad, half of those with an Android device, every single-device Android session and a fixed 35\% of single-device iOS sessions, chosen by a hash of the session identifier and independent of any outcome. The fleet sets keep device-sessions that uploaded at least five depth grids: on iOS, 2,668 of 5,690 without LiDAR and 1,530 of 3,156 with LiDAR, and on Android 2,864 running MoGe-2. The shared set keeps every device in a multi-device session that received a shared-frame solution, whether or not it uploaded depth, so it overlaps the fleet sets only in part. The unit of analysis is the device-session, one device in one session. A user who opened the application several times contributes several device-sessions, and identifiers do not allow clustering by user. Bootstrap intervals resample whole sessions. Table 2 summarizes the data.

\begin{table}[t]
\centering
\small
\captionof{table}{Table 2. Evaluation data.}
\begin{tabular}{>{\raggedright\arraybackslash}p{2.7cm}>{\raggedright\arraybackslash}p{4.4cm}>{\raggedright\arraybackslash}p{3.5cm}>{\raggedright\arraybackslash}p{3.8cm}}
\toprule
\textbf{Set} & \textbf{Devices} & \textbf{Reference} & \textbf{Size} \\
\midrule
Fleet without LiDAR & 20 iPhone and 23 iPad models, MoGe-2 on device & ARKit planes and camera translation & 2,668 device-sessions, 164.7 h, 265,582 depth grids \\
LiDAR fleet & 23 models with LiDAR & the same & 1,530 device-sessions, 81.8 h, 369,309 depth grids \\
Android fleet & 431 models from 18 manufacturers, MoGe-2 on device & ARCore planes and camera translation & 2,864 device-sessions, 129.4 h, 95,985 depth grids \\
Shared & 2 to 6 devices per session & solver state & 4,088 sessions with 7,301 iPhones and iPads; 3,048 Android devices \\
Controlled, sparse & iPhone 11 Pro & ARKit feature points & 36 frames, 3 sessions \\
Controlled, dense & iPhone 15 Pro & LiDAR depth & 4 views, 160,511 pixels \\
Controlled, live & iPhone 15 Pro, MoGe-2 on device & LiDAR depth & 245 frames, 4 sessions \\
\bottomrule
\end{tabular}
\end{table}

Users chose their own rooms and motion, so the fleet sets are not a controlled sample. Telemetry samples depth less often than the model runs, so a device-session contributes a median of 77 depth grids without LiDAR, 206 with it and 22 on Android. LiDAR devices are newer Pro models, and their users may differ in rooms and behaviour as well as in sensors, so the LiDAR fleet is a reference population, not a matched control.

\subsection{Measures}

All fleet measures use the recorded depth grid, which is the deployed output after Equation (4) and the floor snap. In every scale measure, rays pointing more than 0.15 rad (8.6\textdegree{}) below the horizon are excluded, which removes floor pixels that the snap could make agree with ARKit by construction. Shares ``within 10\%'' mean $|\ln x| < \ln 1.1$. On Android, ARCore's planes and camera poses take the place of ARKit's.

\emph{Time to geometry.} The time from session start to the first recorded depth grid and to the first vertical plane reported by ARKit. Because every session lasted at least 60 s, the shares at 10, 30 and 60 s are plain proportions. Sessions that end before an event are censored in the curves beyond 60 s (Kaplan--Meier).

\emph{Scale against wall planes.} For up to 60 depth grids per session, pixels whose rays hit the polygon of an ARKit vertical plane known at that time, between 0.3 m and 5 m, give a ratio of predicted depth to ray--plane depth. Planes must have an area of at least 0.3 m². We discard pixels whose ratio is off by more than a factor of 2 (21\% of hits without LiDAR and 16\% with it, mostly furniture in front of walls), take the median per frame and plane, then the median over the session.

\emph{Scale against camera translation.} ARKit's camera translation is metric. For pairs of depth grids within 4 s whose camera centres are at least 0.3 m apart, we scale the first grid by $k$, move its points into the second camera with the ARKit poses, and compare them with the second grid scaled by $k$. Under pure rotation every $k$ fits equally well. With translation only the correct scale makes the two views agree. We search $k$ over a factor of $e^{\pm 0.8}$, take the value $\hat{k}$ that minimizes the median absolute log-ratio, and report $1/\hat{k}$, the ratio of predicted to metric depth, as the median over up to 60 pairs. Pairs whose best $k$ lies at the edge of the search range are unidentifiable and are dropped.

\emph{Frame-to-frame scale change.} For consecutive grids at most 2.5 s apart whose camera centres moved less than 10 cm, the median ratio $r$ of reprojected to observed depth, reported as $|\ln r|$ in percent. With a small baseline a constant scale error contributes little, so this measures scale that changes over time, plus pose and sensor error. Pairs in which the same model output was recorded twice ($|\ln r| < 10^{-4}$) are excluded.

\emph{Wall thickness.} For each ARKit wall plane, the perpendicular offsets of all depth points that hit it within 5 m, pooled over the session. Pixels whose depth is off from the plane by more than a factor of 2 are discarded, as in the wall-scale measure. We report the interquartile range of the offsets, as the median over planes with at least 200 points from at least three frames. It grows with scale error, scale change and the error of the plane itself.

\emph{Shared frame.} Whether and when each device's state became converged, the predicted lateral error $\sigma$ of Section 3.1 while converged, and the change in yaw after convergence, taken as the median absolute difference between each converged solution's yaw and the yaw at first convergence. Sightings are counted in three classes by their range: \emph{precise} ($\sigma_r \le 0.12\,\mathrm{m} + 0.05\,d$), \emph{coarse} ($\sigma_r \le 0.35\,d$) and \emph{bearing only}. The distance of a sighting is the horizontal distance from the observing camera to the observed user's floor point.

\subsection{Controlled study}

On four views recorded with an iPhone 15 Pro, we compared three models on the same frames against LiDAR (0.35 m to 5 m). MoGe-2 ran offline at full precision with its own field-of-view estimate, which differs from the deployed configuration. Depth Anything V2 Metric Small ran at $252 \times 196$, the input that fitted the time budget of older phones, and at $518 \times 392$. ZipDepth was fitted to half of the LiDAR pixels and scored on the other half, which gives it an advantage that no deployed system has. On 36 frames from three iPhone 11 Pro sessions, we also scored the models against ARKit feature points. In a third test, an iPhone 15 Pro ran the deployed MoGe-2 model on its Neural Engine every 1.5 s during use, and the raw output, before Equation (4), was scored against that frame's LiDAR depth.

\subsection{Offline correction}

To estimate how much of the scale error and of the frame drift could be removed using only the recorded data, we reprocessed a sample of sessions offline. Note that none of the procedures described in this section is part of the deployed system.

\emph{Sample.} From the study period we selected, in the order of a hash of the session identifier, 70 iOS and 70 Android device-sessions without LiDAR that had recorded at least 40 depth grids (30 on Android) and for which a wall-thickness value was available in the fleet analysis, as well as 66 sessions in which a LiDAR device with at least 100 depth grids and a device running MoGe-2 with at least 40 depth grids (25 on Android) had both converged. For each device, we used the depth grids of its longest tracking segment, subsampled evenly to at most 160 grids.

\emph{Scale per frame.} For each device, we estimate one scale factor per depth grid such that the grids agree in the regions where they overlap. The estimation proceeds in six rounds. In each round, every grid is rescaled by the median ratio between the depth predicted at its pixels by the other grids of the device, after reprojection with the ARKit or ARCore poses, and its own depth. This update is damped by a factor of 0.7, and ratios exceeding a factor of 1.6 are regarded as occlusions and ignored. Finally, the factors are divided by their median, so that the overall scale of the session remains unchanged. The estimation uses neither planes nor LiDAR nor any other device. We additionally consider a \emph{causal} variant, in which each grid is rescaled to agree only with the 40 grids preceding it, as would be required on a phone at run time. We compare the wall thickness (Section 4.2) before and after the correction, on the devices for which it can be measured in all three cases.

\emph{Frame drift.} In sessions in which a LiDAR device mapped the same room, we fuse the grids of the LiDAR device in its own frame into a reference surface and place the points of a second device in three different ways. In the first, each grid is placed using the transform reported by the solver at the time of the grid. In the second, a single rigid transform is used for the entire session, namely the yaw and horizontal translation that minimize the mean distance to the reference, truncated at 0.5 m, found by a search that starts from the reported transform. In the third, the scale factors are applied, together with one session scale derived from the wall planes of the device itself, before the rigid transform is determined. We report the mean distance to the reference surface of the points between 0.15 m and 2.3 m above the floor, as well as the share of points within 10 cm.

\section{Results}

\subsection{Shared frame}

We first evaluate how often and how quickly the shared frame converges. Of the 4,613 iPhones and iPads without LiDAR in 4,088 multi-device sessions, 3,925 (85\%) reached the converged state, compared with 2,340 of 2,688 LiDAR devices (87\%). Among the devices that converged, the median time to convergence was 24 s without LiDAR and 22 s with it, and 72\% and 75\% of the devices, respectively, had converged after 60 s (Figure 4b). In the 1,076 sessions that contained devices of both classes, 87\% of the devices in each class converged.

\begin{figure}[t]
\centering
\includegraphics[width=1.0\linewidth]{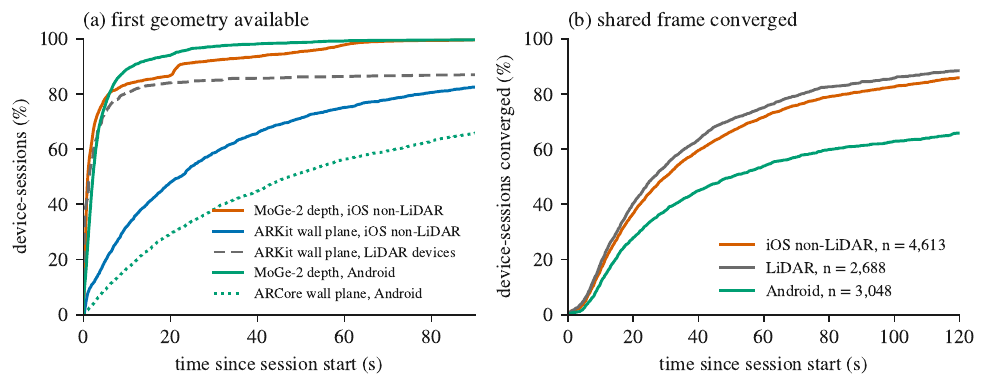}
\captionof{figure}{\textbf{Figure 4.} Deployment results over time. (a) Share of device-sessions with a first MoGe-2 depth grid, and with a first ARKit or ARCore vertical plane, on iPhones and iPads without LiDAR and on Android. The dashed line is the first vertical plane on LiDAR devices. (b) Share of devices whose shared frame had converged. Curves are Kaplan--Meier estimates censored at session end.}
\end{figure}

The two classes converged at similar rates even though their sightings differed considerably. Without LiDAR, 45\% of the sightings had a precise range, 21\% a coarse range and 34\% only a bearing, whereas with LiDAR the corresponding shares were 72\%, 11\% and 17\%. Precise sightings were made at a median distance of 2.1 m and 2.6 m, respectively. A device without a depth sensor thus contributes a larger share of bearings and coarse ranges, which enter the estimation of Section 3.1 with larger standard deviations.

We next examine the contribution of ultra-wideband ranging between iPhones, which was recorded in the sessions of 66\% of these devices. In sessions without it, 81\% of the devices without LiDAR and 82\% of the LiDAR devices converged, compared with 88\% and 89\% in sessions with it. Ultra-wideband ranging thus increases the convergence rate by about seven percentage points, while sightings alone account for most of the result.

While a device was converged, the lateral error predicted by the server had a median of 0.29 m for both classes. However, the yaw continued to change after convergence, by a median of 6.2\textdegree{} without LiDAR and 5.3\textdegree{} with it. Since a change of 6\textdegree{} displaces a point at a distance of 8 m by about 0.8 m, which is more than twice the predicted error, the prediction underestimates the residual motion of the frame. Note that these quantities are reported by the solver itself. In the absence of an external reference such as motion capture, they characterize how often and how quickly the frame converged, but not how accurate it was. Section 5.5 compares the frame with a LiDAR map in one room.

\subsection{Geometry without scanning}

On devices without LiDAR, the first depth grid was recorded after a median of 1.0 s (interquartile range 0.5 to 3.9 s), and by 10 s 84\% of device-sessions had one (Figure 4a). This is an upper bound on when the map had geometry, because telemetry samples depth less often than the model runs. On the same devices ARKit had reported a vertical plane in 32\% of sessions by 10 s, 59\% by 30 s and 75\% by 60 s, and in 364 of the 2,668 sessions it reported none. On LiDAR devices 80\% of sessions had a wall plane within 10 s. Where both events occurred, the plane came a median of 12 s after the first depth grid (2,304 sessions). ARKit tracked a median of 33 feature points per frame and reported 2.9 vertical planes per minute, against 15.7 on LiDAR devices. On Android the first MoGe-2 grid came after a median of 2.2 s, and by 10 s 89\% of device-sessions had one, while ARCore had reported a vertical plane in 17\%.

\subsection{Metric scale}

Table 3 and Figure 5a report the scale of the deployed depth with respect to the two ARKit references. Against wall planes, the median session ratio was 0.987 (95\% interval 0.980 to 0.993) over 1,890 device-sessions, and 50\% of sessions were within 10\%. Against camera translation, the median was 1.05 (1.03 to 1.05) over 2,344 device-sessions, and 34\% were within 10\%. LiDAR devices gave 0.998 and 1.00. On LiDAR devices ARKit's planes are themselves informed by LiDAR, so their small spread is a lower bound on the noise of the measures, and the reference is noisier on devices without LiDAR. Session medians by chip generation from A14 to A19 ranged from 0.96 to 0.99 against wall planes and from 1.02 to 1.04 against translation. The A13 (iPhone 11 family and 9th-generation iPad) read longest, 1.09 and 1.15.

The two estimates agree in the population median and weakly per session (Spearman correlation 0.15, 95\% interval 0.10 to 0.20, over 1,759 sessions with both). Per-session values are still mostly measurement noise, but part of the spread in Table 3 is real variation between sessions. Both references share ARKit's visual-inertial scale, and the deployed output includes the occasional fit to ARKit feature points, so agreement with ARKit is partly by construction. The raw model, measured against LiDAR in Section 5.7, read 16\% short.

\begin{table}[t]
\centering
\small
\captionof{table}{Table 3. Deployed depth against ARKit references (one value per device-session; median, interquartile range in brackets).}
\begin{tabular}{>{\raggedright\arraybackslash}p{5.0cm}>{\raggedright\arraybackslash}p{5.0cm}>{\raggedright\arraybackslash}p{5.0cm}}
\toprule
\textbf{Measure} & \textbf{Without LiDAR, MoGe-2} & \textbf{LiDAR devices} \\
\midrule
Scale against wall planes & 0.987 [0.898 to 1.088], n = 1,890 & 0.998 [0.988 to 1.001], n = 1,281 \\
Sessions within 10\% / 20\% & 50\% / 73\% & 95\% / 98\% \\
Scale against camera translation & 1.05 [0.90 to 1.21], n = 2,344 & 1.00 [0.98 to 1.03], n = 1,435 \\
Sessions within 10\% / 20\% & 34\% / 58\% & 84\% / 91\% \\
Frame-to-frame scale change & 8.3\% [6.0 to 11.9], n = 2,348 & 0.30\% [0.18 to 0.81], n = 1,513 \\
Wall thickness (IQR of offsets) & 31 cm [20 to 47], n = 1,731 & 7.1 cm [4.7 to 11.8], n = 1,273 \\
\bottomrule
\end{tabular}
\end{table}

\begin{figure}[t]
\centering
\includegraphics[width=1.0\linewidth]{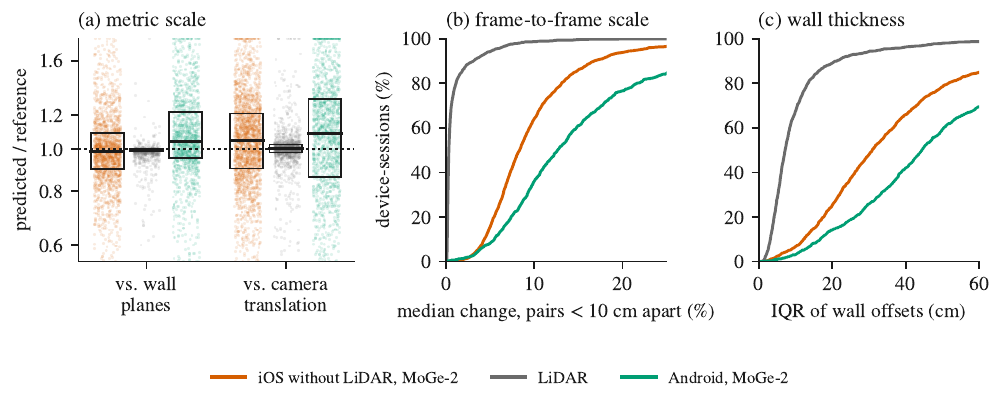}
\captionof{figure}{\textbf{Figure 5.} Scale of the deployed depth. (a) Per-session ratio of predicted depth to wall-plane depth and to depth implied by camera translation; bars are medians and boxes interquartile ranges. (b) Median frame-to-frame scale change per session, for frames less than 10 cm apart. (c) Spread of depth points around wall planes. Orange, iPhones and iPads without LiDAR; green, Android, both running MoGe-2; grey, LiDAR devices. References are ARKit's on iOS and ARCore's on Android.}
\end{figure}

\subsection{Consistency over time}

The scale varied between nearby frames even after the correction of Equation (4). The median frame-to-frame change was 8.3\% per session (95\% interval 8.0\% to 8.5\%), against 0.30\% for LiDAR (Figure 5b). Of 93,134 pairs without LiDAR, 39\% differed by more than 10\% and 11\% by more than 30\%, against 6\% above 10\% for LiDAR. The interquartile range of depth points around ARKit wall planes was 31 cm, about four times the 7.1 cm of LiDAR devices (Figure 5c). This is consistent with walls being widened by scale that changes between the frames that observe them, although error in ARKit's planes also contributes. Figure 6 shows this in two rooms. In each, a LiDAR device and a device running MoGe-2 mapped the same walls in one session, and the walls from MoGe-2 appear thickened or doubled where LiDAR gives a single surface.

\begin{figure}[t]
\centering
\includegraphics[width=1.0\linewidth]{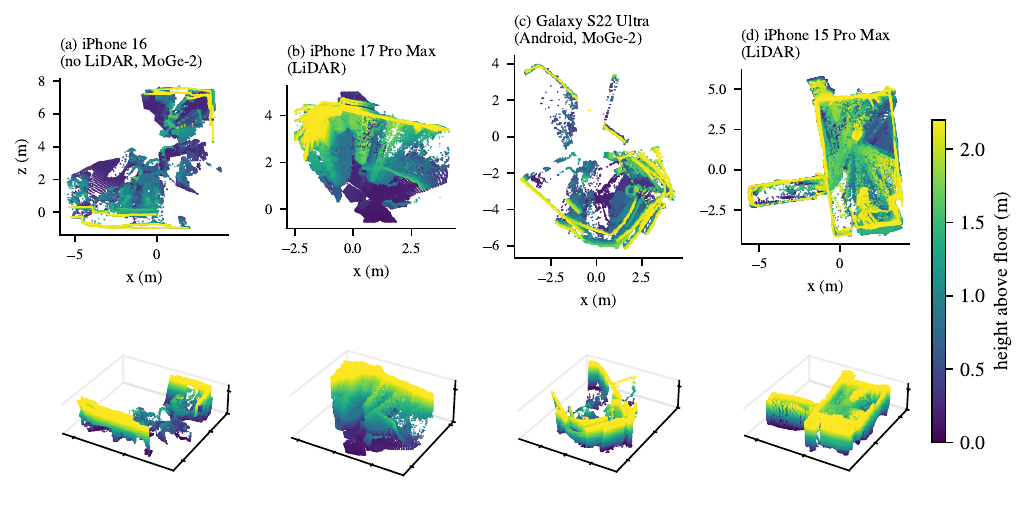}
\captionof{figure}{\textbf{Figure 6.} Deployed depth in two rooms, top-down (top) and in 3D (bottom), coloured by height above the floor. Each device is shown in its own AR frame over its densest 20 s, fused from the recorded depth grids without further processing. (a, b) An iPhone 16 without LiDAR and an iPhone 17 Pro Max with LiDAR in one session. (c, d) A Galaxy S22 Ultra running MoGe-2 and an iPhone 15 Pro Max with LiDAR in another. The sessions were chosen because each device's scale against wall planes was close to the median of its class.}
\end{figure}

\subsection{Offline correction of scale and drift}

The two main sources of error identified above, namely the variation of scale between frames and the continued motion of the shared frame, can both be reduced using data that the devices already record (Section 4.4).

\emph{Scale.} For the 98 iPhones and iPads without LiDAR in the sample, which comprise 17 models, estimating one scale factor per frame reduced the median wall thickness from 27 cm (95\% interval 25 to 32) to 17 cm (14 to 19), and the thickness decreased for 85\% of the devices. The estimated factors deviated from 1 by a median of 13\%. The causal variant, which only has access to earlier frames, yielded 17 cm (15 to 20). For comparison, the LiDAR devices in the same sample had a wall thickness of 7.8 cm (n = 62). For the 70 Android devices, the wall thickness decreased from 50 cm (43 to 59) to 22 cm (20 to 26) with the full estimation, for 90\% of the devices, and to 29 cm (25 to 34) with the causal variant. Since the planes that define the wall thickness are not used in the estimation, this reduction is not obtained by construction. The correction closes about half of the gap to LiDAR on iOS (52\%) and two thirds on Android (66\%). It cannot, however, compensate for shape errors within a frame, for errors in the camera poses or for errors in the planes themselves.

\emph{Drift.} Figure 7 shows a room that was mapped in a single session by a Galaxy S22 Ultra running MoGe-2 and by two LiDAR iPhones, one of which is used as the reference. This is the second room of Figure 6 with the same reference device. It was selected there because the scale of each device was close to the median of its class, before any correction had been computed. When every depth grid is placed using the transform reported by the solver at the corresponding time, the points of the Galaxy lie at a mean distance of 18 cm from the reference surface, and 52\% of them are within 10 cm. A single fixed transform reduces this distance to 11 cm (68\%), and the additional scale correction reduces it to 7 cm (83\%). For the second LiDAR iPhone, a fixed transform reduces the distance from 14 cm to 8 cm. After both corrections, the monocular map thus agrees with the reference approximately as well as the map of a second LiDAR device. The transforms reported for the Galaxy deviated from its fixed transform by a median of 14\textdegree{} in yaw and 0.3 m in position, and those reported for the second iPhone by 4\textdegree{} and 0.2 m. The fixed transforms of the Galaxy obtained before and after the scale correction differed by 4.5\textdegree{}, which indicates the uncertainty of these values.

\begin{figure}[t]
\centering
\includegraphics[width=0.95\linewidth]{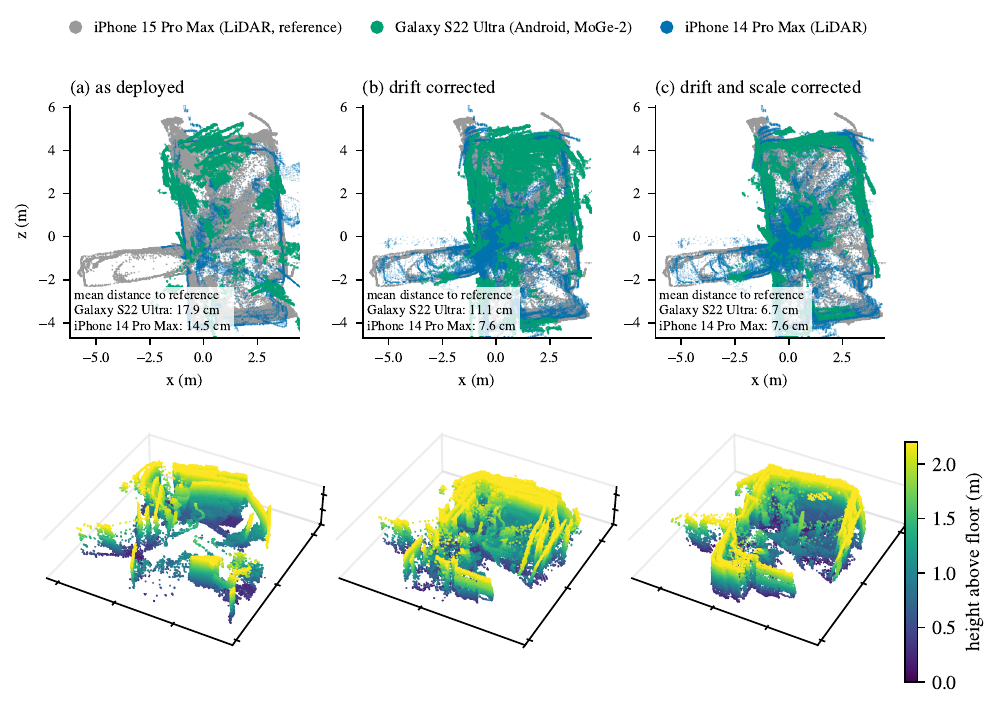}
\captionof{figure}{\textbf{Figure 7.} Offline correction in one room. Top: points between 0.9 m and 1.6 m above the floor, seen from above; grey is the LiDAR reference. (a) Every depth grid placed by the shared-frame transform reported at its time. (b) One fixed rigid transform per device. (c) The same after one scale factor per frame and one session scale for the Galaxy. Bottom: the Galaxy's own points at the same three stages, coloured by height above the floor.}
\end{figure}

We applied the same procedure to the 43 pairs of a MoGe-2 device and a LiDAR device in the sample whose converged periods overlapped. However, the rigid alignment was reliable for only 5 of these pairs, where we consider an alignment reliable if it changes by at most 5\textdegree{} and 0.3 m when the scale correction is applied and at least 85\% of the points lie within 0.5 m of the reference. The main reasons are that players move through the room and that the two devices often observe different parts of it. Given the small number of pairs, we do not report a distribution. Figure 7 should therefore be regarded as a single example, in which the alignment can be verified visually and against the second LiDAR device. The accuracy of the shared frame across the fleet remains unmeasured.

\subsection{Runtime}

The 95th percentile of frame time had a median of 17.3 ms on iPhones and iPads without LiDAR and 17.4 ms on LiDAR devices, close to the 16.7 ms of 60 Hz. Across chip generations from A13 to A19 the median ranged from 17.2 to 18.0 ms. We did not measure the same devices with the model disabled or over long sessions, so this shows only that frame times matched those of LiDAR devices. It does not show that the model costs nothing. Model latency was measured on two phones. An iPhone 11 Pro (A13) took a median of 306 ms per prediction on its CPU (95th percentile 436 ms, 477 predictions), which the timing test of Section 3.2 chose over the Neural Engine. An iPhone 15 Pro took 34.6 ms with the Neural Engine (245 predictions). On an A13 device the map is therefore updated about three times per second.

\subsection{Controlled accuracy}

On the four LiDAR views (Table 4, Figure 8b), MoGe-2 placed 97\% to 100\% of pixels within a factor of 1.25 of LiDAR on three views and 56\% on the fourth, without any fitting. Depth Anything V2 at the $252 \times 196$ input placed under 5\% of pixels within that bound on every view, and at $518 \times 392$ between 5\% and 97\%. ZipDepth, fitted to the LiDAR of each view, reached only 47\% and 68\% on two views, likely because inverting a disparity fit amplifies small errors at far pixels. On the 36 iPhone 11 Pro frames scored against ARKit feature points (Figure 8a), the median per-frame AbsRel was 0.22 for ZipDepth fitted to half of each frame's points, 0.77 for Depth Anything V2 as run on the phone, and 0.23 for MoGe-2. Feature points cover only textured surfaces, which may be why the dense views separate the models better. Four views from one residence cannot establish accuracy in general. On these views, the point-map model was the only one within a factor of 1.25 on most pixels without fitting.

\begin{table}[t]
\centering
\small
\captionof{table}{Table 4. Controlled comparison on four iPhone 15 Pro views against LiDAR.}
\begin{tabular}{>{\raggedright\arraybackslash}p{5.4cm}>{\raggedright\arraybackslash}p{1.6cm}>{\raggedright\arraybackslash}p{5.0cm}>{\raggedright\arraybackslash}p{2.0cm}}
\toprule
\textbf{Model and input} & \textbf{Fitted to LiDAR} & \textbf{Pixels within a factor of 1.25, per view} & \textbf{AbsRel, median over views} \\
\midrule
Depth Anything V2 Metric Small, $252 \times 196$ & no & 4.7\%, 0.2\%, 0.0\%, 0.2\% & 1.28 \\
Depth Anything V2 Metric Small, $518 \times 392$ & no & 97\%, 96\%, 37\%, 5\% & 0.20 \\
ZipDepth, affine fit & yes & 94\%, 47\%, 98\%, 68\% & 1.15 \\
MoGe-2, full precision & no & 56\%, 100\%, 100\%, 97\% & 0.06 \\
\bottomrule
\end{tabular}
\end{table}

\begin{figure}[t]
\centering
\includegraphics[width=1.0\linewidth]{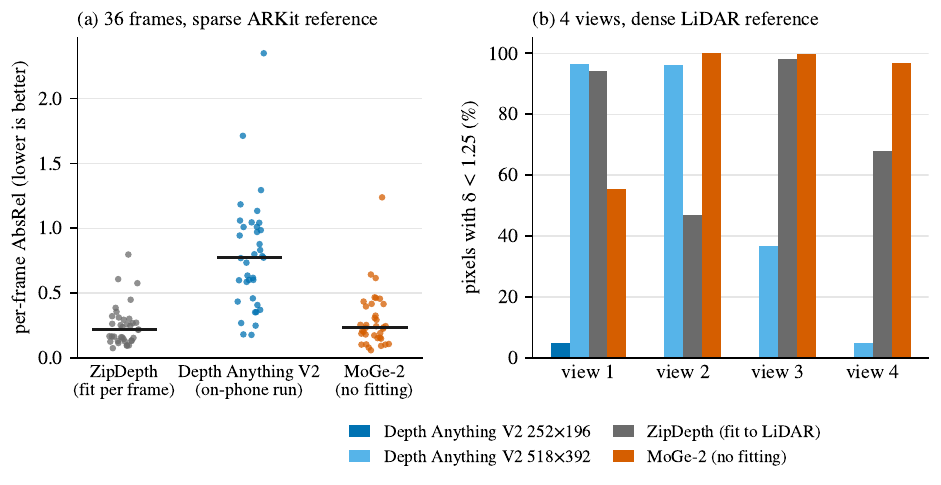}
\captionof{figure}{\textbf{Figure 8.} Controlled comparison. (a) Per-frame AbsRel against ARKit feature points on 36 iPhone 11 Pro frames from three sessions; black bars are medians, and ZipDepth was fitted to half of each frame's points and scored on the other half. (b) Share of pixels within a factor of 1.25 of LiDAR on the four views.}
\end{figure}

The raw on-device output on the iPhone 15 Pro was less accurate (Figure 9). Across 245 frames the per-frame AbsRel had a median of 0.20 and the per-frame scale ratio a median of 0.84, with session medians from 0.68 to 0.99. Consecutive frames 1.5 s apart differed in scale by a median of 12\%. Replaying a temporal version of Equation (4), which pulls each frame toward the mean of the five before it, reduced that to 7\% and changed the session medians by at most 5\%.

\begin{figure}[t]
\centering
\includegraphics[width=1.0\linewidth]{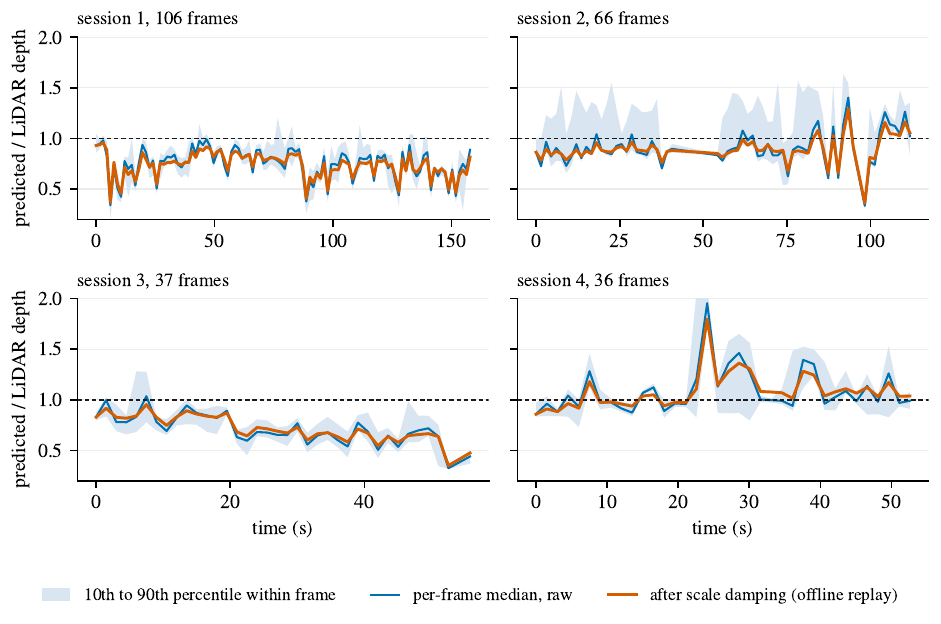}
\captionof{figure}{\textbf{Figure 9.} Raw MoGe-2 depth against LiDAR on an iPhone 15 Pro, one frame every 1.5 s in four sessions. The blue line is each frame's median ratio of predicted to LiDAR depth, the band is the 10th to 90th percentile within the frame, and the orange line replays a temporal version of Equation (4) offline.}
\end{figure}

The raw on-device depth was about 16\% short (median ratio 0.84), while the deployed depth in the fleet was within 2\% of ARKit's wall planes and 5\% of its camera translation. The two differ in reference (LiDAR against ARKit geometry), in rooms and devices (one residence and one phone against many), and in processing (raw output against the deployed output after the floor snap and the occasional fit to feature points). These data do not separate the causes. Running the deployed pipeline on LiDAR devices in the field and comparing it with their LiDAR depth would.

\subsection{Android}

On Android, ARCore Depth exists only on phones that support the Depth API. Of the 710 Android models in our data, 352 did not, and before the pipeline replaced ARCore Depth with MoGe-2 those phones recorded no depth at all (0 of 670 single-device sessions on the previous release, from 19 October). With MoGe-2, 40\% of their single-device sessions recorded depth, and 65\% on phones with the Depth API, against 52\% with ARCore Depth before (Figure 10a). Telemetry does not record depth in every session: LiDAR devices, which always have depth, recorded it in 61\%.

The model was less accurate on Android than on iOS (Figures 5 and 10b). Against ARCore's wall planes the median session ratio was 1.04 (95\% interval 1.03 to 1.06) over 1,375 device-sessions, and 44\% were within 10\%. Against camera translation it was 1.09 (1.07 to 1.10), with 23\% within 10\%. Scale changed by a median of 12.6\% between nearby frames, and depth points around walls had an interquartile range of 44 cm. The scale also depended on the phone: across 66 models with at least ten device-sessions, the median wall-plane ratio ranged from 0.92 to 1.43, and 74\% of models were within 10\% (Figure 10c). ARCore Depth, where it ran, gave 0.99 and 0.995, but it is computed from the same tracking as ARCore's planes and poses and is smoothed over time, so its agreement with them is largely by construction. The smaller input, the coarser output and ARCore's sparser tracking on low-cost phones (a median of 27 feature points per frame) may each contribute; our data cannot separate them.

\begin{figure}[t]
\centering
\includegraphics[width=1.0\linewidth]{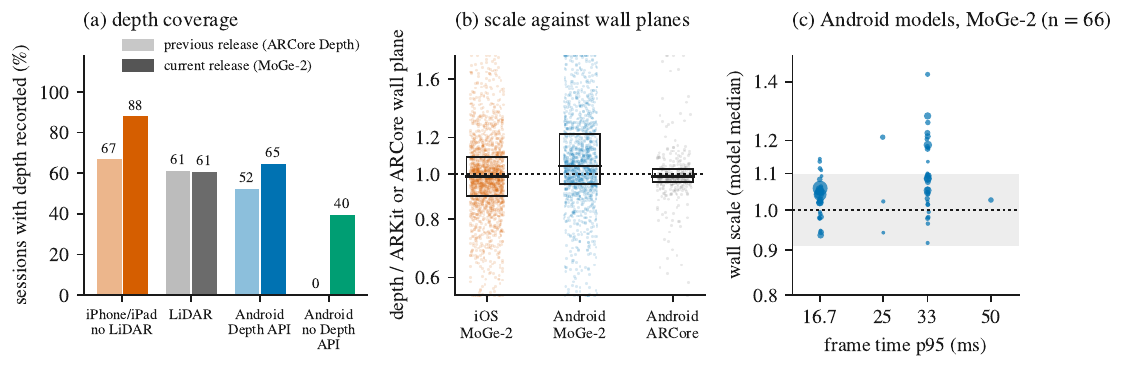}
\captionof{figure}{\textbf{Figure 10.} Android. (a) Share of single-device sessions that recorded depth, by device class, with ARCore Depth on the previous release (light) and with MoGe-2 (dark). (b) Per-session scale against wall planes for MoGe-2 on iOS and Android and for ARCore Depth. (c) Median frame time and median wall-plane scale for each Android model with at least ten device-sessions; marker size is the number of device-sessions.}
\end{figure}

In multi-device sessions, 66\% of 3,048 Android devices converged (Figure 4b), against 75\% of iPhones and iPads without LiDAR and 75\% of LiDAR devices in the same sessions. Only 7\% of Android sightings had a precise range and 24\% a coarse one, and 69\% were bearings only. Android phones also had no ultra-wideband path. Either difference may explain the lower rate. iPhones and iPads without LiDAR converged in 81\% of cases in sessions without ultra-wideband ranging (Section 5.1), which suggests that the sparser ranges matter more.

\section{Discussion and Limitations}

Our results show that mutual sightings are sufficient to register co-located phones in a common frame without a shared map. On iOS, the shared frame converged for 85\% of the devices without LiDAR and 87\% of the LiDAR devices, in a median of 24 s and 22 s, respectively, even though the devices without LiDAR obtained precise ranges for fewer of their sightings. Ultra-wideband ranging increased the convergence rate by about seven percentage points where it was available, while sessions without it still converged in 81\% and 82\% of the cases. On Android, where precise ranges were rare and no radio path existed, 66\% of the devices converged. However, the frame continued to move after convergence. The yaw changed by a median of about 6\textdegree{}, which is more than the error predicted by the solver accounts for. Since each device adopts every new solution immediately, any correction of the estimate displaces all shared content. In the room of Figure 7, a single fixed transform fitted the maps better than the sequence of reported transforms.

Monocular point maps made dense depth available within seconds on devices on which ARKit required tens of seconds to report a wall, or never reported one. Across the iOS fleet, the median scale of the deployed depth was within 2\% of ARKit's wall planes and within 5\% of its camera translation. The largest remaining depth error is the variation of scale between nearby frames, which amounts to a median of 8\% compared with 0.3\% for LiDAR and is consistent with depth points spreading about four times as widely around walls. The frame-to-map alignment of Equation (4) reduces this variation but does not eliminate it, and it cannot correct a session whose overall scale is wrong, since it preserves the mean scale of the session. On A13 devices, this appears as a scale that is about 10 to 15\% too long. On Android, the model provided depth on phones without the Depth API, but its scale was less accurate and varied with the phone model.

The offline correction indicates how much of this error is recoverable. Estimating one scale factor per frame from the overlapping frames of the device itself reduced the wall thickness from 27 cm to 17 cm on iOS and from 50 cm to 22 cm on Android, and the causal variant reached 17 cm and 29 cm. Both variants rely only on the frames and poses of the device itself. The remaining gap to LiDAR (8 cm) cannot be explained by a per-frame scale.

\emph{Implications for multi-user AR.} The main practical consequence is that a multi-user session does not need a setup phase. With map-based registration, one user scans the room and the others must then view the same surfaces before any content can be shared. With mutual sightings, registration takes place while the application is already in use, requires no action from the users, and admits a new device as soon as it comes into the view of a device that is already registered. A second consequence concerns privacy and bandwidth. For the purpose of registration, the devices exchange only their poses and the bearings and ranges of the people they see. Neither images nor a map of the room leave the device, and a sighting consists of a few numbers that are sent at most three times per second. This makes the method suitable for applications that are used in private rooms or over poor connections. Finally, the method does not depend on a depth sensor, and in our deployment it ran on iPhones without LiDAR and on Android phones without the Depth API.

\emph{Other platforms.} The method makes two assumptions about a device: that it tracks its own motion in a gravity-aligned frame, and that it can detect the carrier of another device in its camera image. Head-worn displays satisfy both. Since people who talk to each other also look at each other, we expect sightings to be more frequent on glasses than on handheld phones, and glasses without depth sensors would benefit in the same way as the phones studied here. The same assumptions hold for a phone and a robot, and for groups of robots or drones with visual-inertial odometry, where a detector for the other robot takes the place of the person detector. This is the setting of the mutual localization literature \cite{r22,r23,r24}, and our results suggest that coarse and intermittent observations are sufficient in practice. A further case is a team that works in a building for which no map exists, such as firefighters with body-worn cameras, who cannot be asked to scan a room together before they start.

\emph{Shared reconstruction.} Once several devices share a frame, their individual maps can be combined into one. Figure 7 gives an example with three devices in one room, in which the map of a phone without a depth sensor agrees with the maps of two LiDAR devices after the corrections of Section 5.5. This suggests that a room could be reconstructed jointly by the people who are in it, with each device contributing the part it observes and a LiDAR device, where one is present, providing a scale reference for the others. In such a system, the frame obtained from sightings would serve as the initial alignment and the overlap between the maps would refine it, which would also address the drift reported in Section 5.1.

\emph{Future work.} Three extensions follow directly from our findings. First, the scale correction should run on the device. The causal variant, which uses only earlier frames, already reduced the wall thickness to 17 cm on iOS and 29 cm on Android, so that much of the benefit does not depend on later frames. Second, the shared frame should be held, or changed only gradually, once it has converged, since every correction is currently applied at once. Third, the accuracy of the shared frame should be measured against an external reference such as motion capture, and the method should be evaluated in applications in which users face each other less often than in a game. On Android, calibrating the depth scale per device model is a further step.

Our evaluation has the following limitations.

\begin{itemize}[leftmargin=1.6em,itemsep=2pt]
\item The fleet measures are proxies derived from ARKit or ARCore, which share their visual-inertial scale with both scale references, and the deployed output is partly fitted to them. The two scale estimates agree only weakly per session, so they mainly support statements about the population.
\item The telemetry stores depth after the deployed corrections, so it cannot separate the raw model from Equation (4) and the floor snap in the field.
\item The data cover 16 days and the users of one application. On Android the references are ARCore's, and ARCore's tracking is weaker on low-cost phones. Users chose rooms and motion, only sessions of at least 60 s were kept, and one user can contribute several device-sessions. The LiDAR fleet differs from the fleet without LiDAR in devices and users as well as in its sensor.
\item Query correctness for occlusion, collision and placement was not evaluated in the field.
\item The shared-frame results use the solver's own convergence criterion. Accuracy was not measured against an external reference, and the comparison with a LiDAR map is reported for one room. The contributions of the compass prior and of ultra-wideband ranging were not separated beyond the split by sessions in Section 5.1.
\item The offline correction was run on a sample of 168 devices, and wall thickness depends on ARKit's or ARCore's planes. Its effect on occlusion and placement was not tested.
\item Ground-truth accuracy comes from one residence and one LiDAR phone, and the dense comparison did not use the deployed model configuration.
\end{itemize}

The method itself has limitations as well. Users must come into each other's view at some point, each sighting must be attributed to the correct user, and the reduction to three unknowns assumes a single floor level. Furthermore, the deployed application is a game in which players aim at one another, so that sightings are frequent. In applications in which users rarely face each other, the frame can be expected to converge later.

\section{Conclusion}

We presented a method for registering co-located phones in a common coordinate frame from brief sightings of each other's users. A single pair of mutual bearings determines the relative yaw, and ranged sightings determine the translation, without a shared visual map and without a scanning step. In a deployed application, 85\% of 4,613 iPhones and iPads without LiDAR and 66\% of 3,048 Android phones reached a converged frame, the former in a median of 24 s. In addition, a metric point-map network with intrinsics-based shift recovery provided the same devices with dense depth within seconds. On iOS, its median scale was within 2\% of ARKit's wall planes and within 5\% of its camera translation, and frame times matched those of LiDAR devices on chips from A13 to A19. Two weaknesses remain. First, the shared frame continues to move after convergence. Second, the depth scale varies from frame to frame, which thickens surfaces in the map, and on the slowest chips and on Android the scale of an entire session can additionally be off. By reprocessing recorded sessions, we showed that a per-frame scale estimated from the frames of the device itself reduces the wall thickness from 27 cm to 17 cm on iOS and from 50 cm to 22 cm on Android, and that, in one room, a fixed transform combined with this correction brought a monocular map as close to a LiDAR map as that of a second LiDAR device. In future work, we plan to run such a correction on the device, to hold or rate-limit the frame after convergence, to calibrate the scale per device model on Android, and to measure the accuracy of the shared frame against an external reference.


\begin{thebibliography}{99}

\bibitem{r1} Apple. \emph{ARKit}. Apple Developer Documentation, \url{https://developer.apple.com/documentation/arkit}.

\bibitem{r2} Google. \emph{Depth API}. ARCore Developer Guides, \url{https://developers.google.com/ar/develop/depth}.

\bibitem{r3} J. Valentin et al. ``Depth from motion for smartphone AR.'' \emph{ACM Transactions on Graphics (SIGGRAPH Asia)}, 37(6), 193, 2018.

\bibitem{r4} R. Wang et al. ``MoGe-2: Accurate monocular geometry with metric scale and sharp details.'' \emph{Advances in Neural Information Processing Systems (NeurIPS)}, 2025. \href{https://arxiv.org/abs/2507.02546}{arXiv:2507.02546}.

\bibitem{r5} R. Du et al. ``DepthLab: Real-time 3D interaction with depth maps for mobile augmented reality.'' \emph{ACM Symposium on User Interface Software and Technology (UIST)}, 2020, pp. 829--843.

\bibitem{r6} P. Ondr\'{u}\v{s}ka, P. Kohli, and S. Izadi. ``MobileFusion: Real-time volumetric surface reconstruction and dense tracking on mobile phones.'' \emph{IEEE Transactions on Visualization and Computer Graphics}, 21(11), 1251--1258, 2015.

\bibitem{r7} F. Tosi, L. Bartolomei, M. Poggi, and S. Mattoccia. ``ZipDepth: Bringing lightweight zero-shot monocular depth anywhere, on any device.'' 2026. \href{https://arxiv.org/abs/2607.08771}{arXiv:2607.08771}.

\bibitem{r8} L. Yang et al. ``Depth Anything V2.'' \emph{Advances in Neural Information Processing Systems (NeurIPS)}, 2024.

\bibitem{r9} W. Yin et al. ``Metric3D: Towards zero-shot metric 3D prediction from a single image.'' \emph{IEEE/CVF International Conference on Computer Vision (ICCV)}, 2023, pp. 9043--9053.

\bibitem{r10} L. Piccinelli et al. ``UniDepth: Universal monocular metric depth estimation.'' \emph{IEEE/CVF Conference on Computer Vision and Pattern Recognition (CVPR)}, 2024.

\bibitem{r11} A. Bochkovskii et al. ``Depth Pro: Sharp monocular metric depth in less than a second.'' \emph{International Conference on Learning Representations (ICLR)}, 2025. \href{https://arxiv.org/abs/2410.02073}{arXiv:2410.02073}.

\bibitem{r12} R. Wang, S. Xu, C. Dai, J. Xiang, Y. Deng, X. Tong, and J. Yang. ``MoGe: Unlocking accurate monocular geometry estimation for open-domain images with optimal training supervision.'' \emph{IEEE/CVF Conference on Computer Vision and Pattern Recognition (CVPR)}, 2025, pp. 5261--5271.

\bibitem{r13} X. Luo, J.-B. Huang, R. Szeliski, K. Matzen, and J. Kopf. ``Consistent video depth estimation.'' \emph{ACM Transactions on Graphics (SIGGRAPH)}, 39(4), 71, 2020.

\bibitem{r14} R. A. Newcombe et al. ``KinectFusion: Real-time dense surface mapping and tracking.'' \emph{IEEE International Symposium on Mixed and Augmented Reality (ISMAR)}, 2011, pp. 127--136.

\bibitem{r15} H. Oleynikova, Z. Taylor, M. Fehr, R. Siegwart, and J. Nieto. ``Voxblox: Incremental 3D Euclidean signed distance fields for on-board MAV planning.'' \emph{IEEE/RSJ International Conference on Intelligent Robots and Systems (IROS)}, 2017.

\bibitem{r16} K. Tateno, F. Tombari, I. Laina, and N. Navab. ``CNN-SLAM: Real-time dense monocular SLAM with learned depth prediction.'' \emph{IEEE/CVF Conference on Computer Vision and Pattern Recognition (CVPR)}, 2017.

\bibitem{r17} Z. Teed and J. Deng. ``DROID-SLAM: Deep visual SLAM for monocular, stereo, and RGB-D cameras.'' \emph{Advances in Neural Information Processing Systems (NeurIPS)}, 2021.

\bibitem{r18} T. Qin, P. Li, and S. Shen. ``VINS-Mono: A robust and versatile monocular visual-inertial state estimator.'' \emph{IEEE Transactions on Robotics}, 34(4), 1004--1020, 2018.

\bibitem{r19} Apple. \emph{Creating a collaborative session}. Apple Developer Documentation, \url{https://developer.apple.com/documentation/arkit/creating-a-collaborative-session}.

\bibitem{r20} Google. \emph{Cloud Anchors allow different users to share AR experiences}. ARCore Developer Guides, \url{https://developers.google.com/ar/develop/cloud-anchors}.

\bibitem{r21} P. Schmuck and M. Chli. ``CCM-SLAM: Robust and efficient centralized collaborative monocular simultaneous localization and mapping for robotic teams.'' \emph{Journal of Field Robotics}, 36(4), 763--781, 2019.

\bibitem{r22} A. Franchi, G. Oriolo, and P. Stegagno. ``Mutual localization in multi-robot systems using anonymous relative measurements.'' \emph{The International Journal of Robotics Research}, 32(11), 1302--1322, 2013.

\bibitem{r23} X. S. Zhou and S. I. Roumeliotis. ``Determining 3-D relative transformations for any combination of range and bearing measurements.'' \emph{IEEE Transactions on Robotics}, 29(2), 458--474, 2013.

\bibitem{r24} V. Dhiman, J. Ryde, and J. J. Corso. ``Mutual localization: Two camera relative 6-DOF pose estimation from reciprocal fiducial observation.'' \emph{IEEE/RSJ International Conference on Intelligent Robots and Systems (IROS)}, 2013, pp. 1347--1354.

\end{thebibliography}
\end{document}